\pdfoutput=1

\documentclass[11pt]{article}

\usepackage[final]{acl}

\usepackage{times}
\usepackage{latexsym}

\usepackage[T1]{fontenc}

\usepackage[utf8]{inputenc}

\usepackage{microtype}

\usepackage{inconsolata}

\usepackage{graphicx}
\usepackage{xcolor}
\definecolor{deepgreen}{rgb}{0.0, 0.5, 0.0} 
\usepackage{algorithm}
\usepackage{algpseudocode}
\usepackage{graphicx}
\usepackage{wrapfig}
\usepackage{amsmath}
\usepackage{float} 
\usepackage{graphicx} 
\usepackage{pifont}
\usepackage[utf8]{inputenc} 
\usepackage[T1]{fontenc}    
\usepackage{url}            
\usepackage{booktabs}       
\usepackage{amsfonts}       
\usepackage{nicefrac}       
\usepackage{microtype}      
\usepackage{xcolor}         
\usepackage{tikz}
\usepackage{hyperref}
\usepackage{enumitem}
\setlist[itemize]{noitemsep}
\setlist[enumerate]{noitemsep}

\usepackage{booktabs}

\title{Teaching Embodied Reinforcement Learning Agents:\\Informativeness and Diversity of Language Use}

\author{
  Jiajun Xi\thanks{Equal contribution.} 
  \quad Yinong He$^{*}$  
  \quad \textbf{Jianing Yang} 
  \quad \textbf{Yinpei Dai}
  \quad \textbf{Joyce Chai} \\ \\
  University of Michigan \\ \\
 \texttt{\{jiajunxi, heyinong, jianingy, daiyp, chaijy\}@umich.edu}
}

\begin{document}
\maketitle
\newcommand{\jjx}[1]{{\color{blue}#1}}

\begin{abstract}
In real-world scenarios, it is desirable for embodied agents to have the ability to leverage human language to gain explicit or implicit knowledge for learning tasks. Despite recent progress, most previous approaches adopt simple low-level instructions as language inputs, which may not reflect natural human communication. It's not clear how to incorporate rich language use to facilitate task learning.
To address this question,  this paper studies different types of language inputs in facilitating reinforcement learning (RL) embodied agents. More specifically, we examine how different levels of language \textbf{informativeness} (i.e., feedback on past behaviors and future guidance) and \textbf{diversity} (i.e., variation of language expressions) impact agent learning and inference.  
Our empirical results based on four RL benchmarks demonstrate that agents trained with diverse and informative language feedback can achieve enhanced generalization and fast adaptation to new tasks. These findings highlight the pivotal role of language use in teaching embodied agents new tasks in an open world. \footnote{Source code available at \url{https://github.com/sled-group/Teachable_RL}.}
\end{abstract}

\section{Introduction}
Developing embodied agents that can understand and communicate with humans in natural language to learn and accomplish tasks is a long-standing goal in artificial intelligence. In recent years, the integration of human language and reinforcement learning (RL) has seen significant advancements. Unlike traditional RL methods that typically rely on numerical reward signals to guide agent learning, recent works \cite{cheng2023llf, lin2023learning} explore using language as an intuitive and  useful signal to shape an agent's behaviors. For example, when the agent is making mistakes during the task completion, providing language feedback can largely improve the instantaneous performance thus enhancing the overall agent learning efficiency and effectiveness \cite{mccallum2023feedback}. 

However, existing methods generally employ simple instructions, such as \textit{"turn left"} and \textit{"put the apple to the table"} to teach/control an agent \cite{hanjie21grounding, zhang2021-hierarchical, lin2023learning, mccallum2023feedback, ALFWorld20}. While useful, these instructions may not fully reflect the flexibility of language use in task learning and collaboration \cite{chai2018, chai2019teaching, zhang-etal-2022-danli, zhang2023seagull, dai2024racer}. In the real world, humans often express complex language instructions that are more {\bf \em informative}. For instance, when a student makes a mistake, a teacher may help them to retrospect on what went wrong (i.e., \textit{hindsight instructions}) and then guide them on what should be done next to finish the goal (\textit{i.e., foresight instructions}). In addition, humans are likely to engage in conversations with more {\bf \em  diverse} language patterns, describing the same goal with different expressions and styles. Therefore, we ask the following question:

\begin{center}
    \textit{How do the informativeness and diversity of natural language used during RL training affect an agent's ability to learn tasks?}
\end{center}

We take a popular offline RL model - decision transformer (DT) \cite{chen2021decision} - as a backbone architecture and conduct a comprehensive study to examine how informativeness and diversity of language use may impact agents' learning ability. 
To control informativeness, we leverage expert agents' actions as a reference to generate \emph{hindsight} reflection and \emph{foresight} guidance, using hand-crafted language templates.
To increase diversity, we construct a GPT-augmented language pool, where GPT-4 \cite{openai2024gpt4} is used to augment hand-crafted templates into much more natural and richer expressions. We further extended DT into a multi-modal Language-Teachable DT (LTDT) and demonstrated that LTDT agents that are trained with diverse and informative language significantly outperform the counterpart agents that are trained either with simple language alone or with no language inputs. 
Notably, we found that even with just one language template, combining hindsight and foresight feedback together improves agents' performance by an average of 9.86 points (from 37.95\% to 47.81\%) on four popular offline RL benchmarks compared to agents trained without language. When more language diversity is incorporated into training, an additional 10.14 points (from 47.81\% to 57.95\%)  are obtained.

The contributions of this paper can be summarized as follows:
\vspace{-5pt}
\begin{itemize}
   \setlength\itemsep{0pt} 
    \setlength\topsep{0pt} 
    \setlength\parsep{0pt} 
    \item We investigate in detail, for the first time, how language informativeness and diversity affect  
     offline RL agents in task learning, and demonstrate their important roles in improving agents' performance, adaptability, and robustness.
    \item We show that training agents with informative and diverse instructions can intrinsically improve the agent's understanding of the task and lead to better performance.
    \item We propose a simple framework to generate both hindsight and foresight language feedback and enrich language variation without any human annotators. 
\end{itemize}

\section{Related Work}
\vspace{-0.2em}
\paragraph{Offline Reinforcement Learning}
Offline reinforcement learning (RL) has become a focal point of research due to its ability to utilize pre-existing datasets for training agents without real-time interactions. Several algorithms address the unique challenges of offline RL, such as mitigating extrapolation errors and ensuring robust policy evaluation. A survey by \citet{prudencio2023survey} outlines the field's taxonomy and open problems. Benchmarking efforts by \citet{fujimoto2019benchmarking} assess various batch deep RL algorithms. Key approaches include Conservative Q-Learning (CQL) \cite{kumar2020conservative}, Implicit Q-Learning (IQL) \cite{kostrikov2021offline}, and the Decision Transformer (DT) \cite{chen2021decision}, which treats RL as a sequence modeling problem \cite{janner2021offline}. Recent work also explores generalization across tasks \cite{lee2022multi, reed2022generalist, schubert2023generalist}, the use of exploratory data \cite{yarats2022don}, and integrating large language models (LLMs) \cite{mirchandani2023large}. Efficient online RL leveraging offline data is also a focus \cite{ball2023efficient, modhe2023exploiting}.
Our research builds on the Decision Transformer (DT) by integrating language feedback, creating the Language-Teachable Decision Transformer (LTDT). This novel approach incorporates rich, human-like language instructions, improving agent learning through enhanced informativeness and diversity of language inputs.

\paragraph{Language in Reinforcement Learning}
The intersection of natural language and RL offers new ways to develop intuitive and effective learning paradigms for embodied agents. Initial works utilized language for feedback and task instructions \cite{she2017,nguyen2017reinforcement, shridhar2020alfred}. Recent studies have explored various methods for incorporating language feedback in RL, such as the LTC paradigm \cite{wang2023adapting}, lifelong robot learning with human-assisted language planners \cite{parakh2023lifelong}, and frameworks for rich information requests \cite{dai2020learning, tseng-etal-2021-transferable, nguyen2022framework}. Language for corrections \cite{sharma2022correcting, liu2023reflect} and as reward signals \cite{xie2023text2reward, goyal2019using, yu2023language} has shown to enhance agent performance. Vision-language joint training approaches, like CLIP \cite{radford2021learning}, BLIP-2 \cite{li2023blip2}, and InstructBLIP \cite{dai2023instructblip}, demonstrate the potential of combining visual and language modalities for RL tasks \cite{ma2023liv, nguyen2019vision, khandelwal2022simple}. Further, multimodal prompts for robotic manipulation \cite{jiang2023vima, fan2022minedojo} and LLMs for planning in robotics \cite{saycan2022arxiv, huang2022inner, singh2023progprompt, yao2022react, dai2024think} highlight the evolving role of language in RL. Other works, like \cite{mehta2023improving}, focus on generating problem-specific language feedback templates.
In contrast, our work focuses on the informativeness and diversity of language instructions, two problem-agnostic yet easy-to-implement properties. By using both hindsight and foresight language templates and enhancing diversity through GPT-4, we demonstrate notable improvements in agent performance and generalizability, showcasing the impact of complex language inputs in offline RL training.

\begin{figure*}[ht]
    \centering
    \includegraphics[width=0.9\textwidth]{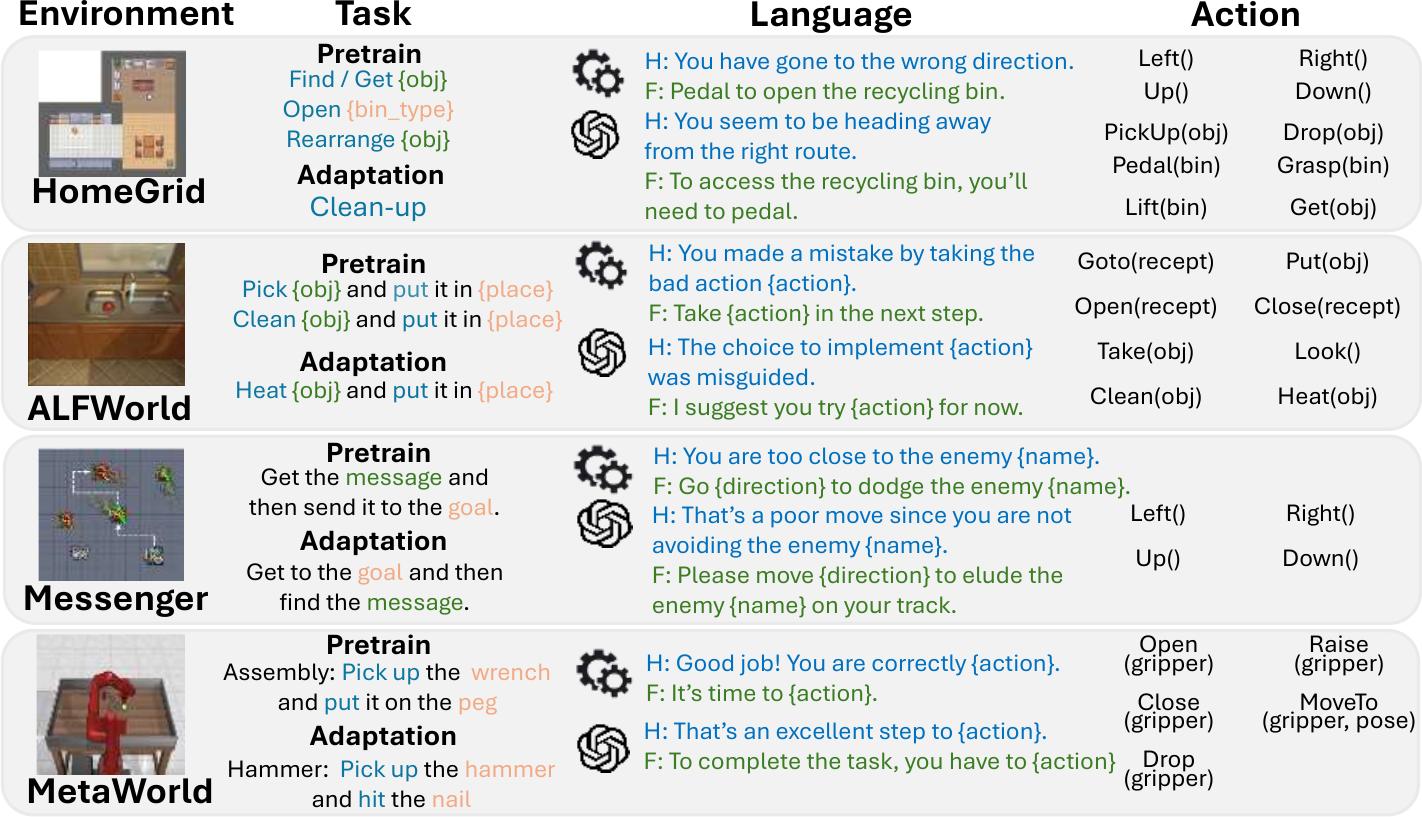}
    \caption{An overview of four environments used for experiments. It shows tasks to be learned in each environment; examples of hindsight (marked \texttt{H}) and foresight (\texttt{F}) language feedback (next to the gear icon are hand-crafted templates and next to the GPT icon are GPT-4 generated feedback); as well as low-level actions in each environment. }
    \label{fig:env_overview}
    \vspace{-15pt}
\end{figure*}
\section{Problem Setting}

In this section, we outline the problem setting by defining the offline reinforcement learning problem (Sec. \ref{sec:def_offline_RL}), and a taxonomy of language feedback (Sec. \ref{sec:tax_lf}). Then we describe the instantiation of such definitions in four different RL environments we used for experiments (Sec. \ref{sec:envs}).

\subsection{Offline Reinforcement Learning}
\label{sec:def_offline_RL}
To support a systematic study of language use, we formulate the problem in the offline reinforcement learning (RL) setting.  
At each time step \( t \), the agent receives an observation \( o_t \), a reward \( r_t \), and a language feedback \( l_t \) for its previous action. The agent then executes an action \( a_t \) according to a policy \(\pi\), which is conditioned on the entire interaction history \( h_t \) up to time \( t \), i.e., $\pi(a_t \mid h_t)$, where \( h_t = \{o_{\leq t}, r_{\leq t}, l_{\leq t}, a_{<t}\} \) represents the history of observations, rewards, language feedback, and past actions up to time \( t \). The agent's goal is to complete the task by maximizing the expected discounted sum of rewards $\mathbb{E}[\sum_{t=1}^{T}\gamma^{t}r_t]$ 
where $T$ is the episode length, and $\gamma$ is the discount factor. 
In offline RL, the training trajectories are pre-collected with an expert agent (a well-trained agent or a planner-based expert with privileged information). The trained agents are evaluated interactively with the environment.  

\subsection{Language Feedback: Informativeness and Diversity}
\label{sec:tax_lf}
We aim to investigate how the \textbf{\em informativeness} and \textbf{\em diversity} of language instructions used during the training of an offline RL agent affect the agent's performance on seen tasks and adaptation to unseen tasks.

\subsubsection{Informativeness}
Informativeness refers to the richness of information content in language feedback. Following \citet{cheng2023llf}, we categorize feedback into two types: \textbf{\em hindsight} and \textbf{\em foresight}. Hindsight feedback involves comments or critiques about the agent's past actions. For example, \textit{"Excellent, you are moving towards the goal!"} encourages the agent to continue its current path, while \textit{"You are getting too close to the enemy."} alerts the agent about a mistake. Hindsight feedback reflects on incorrect actions taken in previous steps, which can guide agents toward success by narrowing down the search space for correct actions (See Appendix \ref{sec:hindsight-impact} for more analysis). Conversely, foresight feedback guides potential future actions. For instance, \textit{"You should go right to get closer to the target."} directs the agent towards the goal, and \textit{"You should go left to avoid the enemy on the right."} helps the agent make strategic decisions to avoid threats. Language feedback is considered most informative when it includes both hindsight and foresight elements, and least informative when neither is present.

\subsubsection{Diversity}
Diversity in language feedback refers to the variety of ways the same information is conveyed. If feedback is provided using only one template, it is less diverse. It becomes more diverse when the same information is expressed in many different ways. The goal is to expose the RL agent to various expressions of the same feedback to enhance its ability to generalize.

\subsection{Environments}
\label{sec:envs}
As shown in Figure \ref{fig:env_overview}, we conduct experiments across four environments—HomeGrid, ALFWorld, Messenger, and MetaWorld—each featuring discrete action spaces, with hand-crafted hindsight and foresight language instructions. 
More information and examples of languages for each environment can be found in Appendix \ref{Appendix:Language_feedback}.  

\noindent\textbf{HomeGrid} \cite{lin2023learning} is a multitask grid world designed to evaluate how well agents can understand and use various types of language to complete tasks. It includes five task types (\textsc{find, get, clean up, rearrange, open}), involving interaction with objects and trash bins with a total of 38 tasks. 
The agent receives a reward of 1 when the task is completed and receives a reward of 0.5 if a subgoal is completed.

\noindent \textbf{ALFWorld} \cite{ALFWorld20} is a text-game environment that aligns with the embodied ALFRED benchmark \cite{shridhar2020alfred} and provides simulation for household tasks. It includes six types of tasks which require the agent to navigate and interact with household objects by following language instructions. The agent gets a reward of 1 when the task is completed. We adopt the hindsight and foresight language templates from LLF-ALFWorld introduced in \cite{cheng2023llf}, which adds an extra language wrapper to the original ALFWorld environment. 

\noindent \textbf{Messenger} \cite{hanjie21grounding} is a grid world with several entities. The agent's task is to retrieve a message from one entity and deliver it to another goal entity, while avoiding enemies. At the start of each episode, the agent is provided with a manual describing the randomized roles of the entities and their movement dynamics. The agent receives a reward of 1 when the task is completed.

\noindent \textbf{MetaWorld} \cite{yu2019meta} is a benchmark that consists of a variety of manipulation tasks performed by a simulated Sawyer robot arm. It includes 50 types of common robot manipulation tasks. We select two of them in our experiments: \textsc{assembly} and \textsc{hammer}. The agent receives a reward of 1 when completing a task.

\section{Data Generation}
\vspace{-0.7em}
To train an agent that can understand language feedback in an offline reinforcement learning manner, we construct an offline dataset $\mathcal{D}$ consisting of two parts:
\vspace{-10pt}
\begin{itemize}
    \item Agent trajectory consisting of task description $Td$ and the tuples $(\hat{R}_t, s_t, a_t)$, where $\hat{R}_t$ represents the reward, $s_t$ is the state, and $a_t$ is the action.
    \item language feedback $l_t$ conveying hindsight and foresight information at each time step.
\end{itemize}
Algorithm \ref{alg:data_collection} outlines the data generation process, and we explain the algorithm in detail in the following sections. 
\begin{figure*}[ht]
    \centering
    \includegraphics[width=0.93\textwidth]{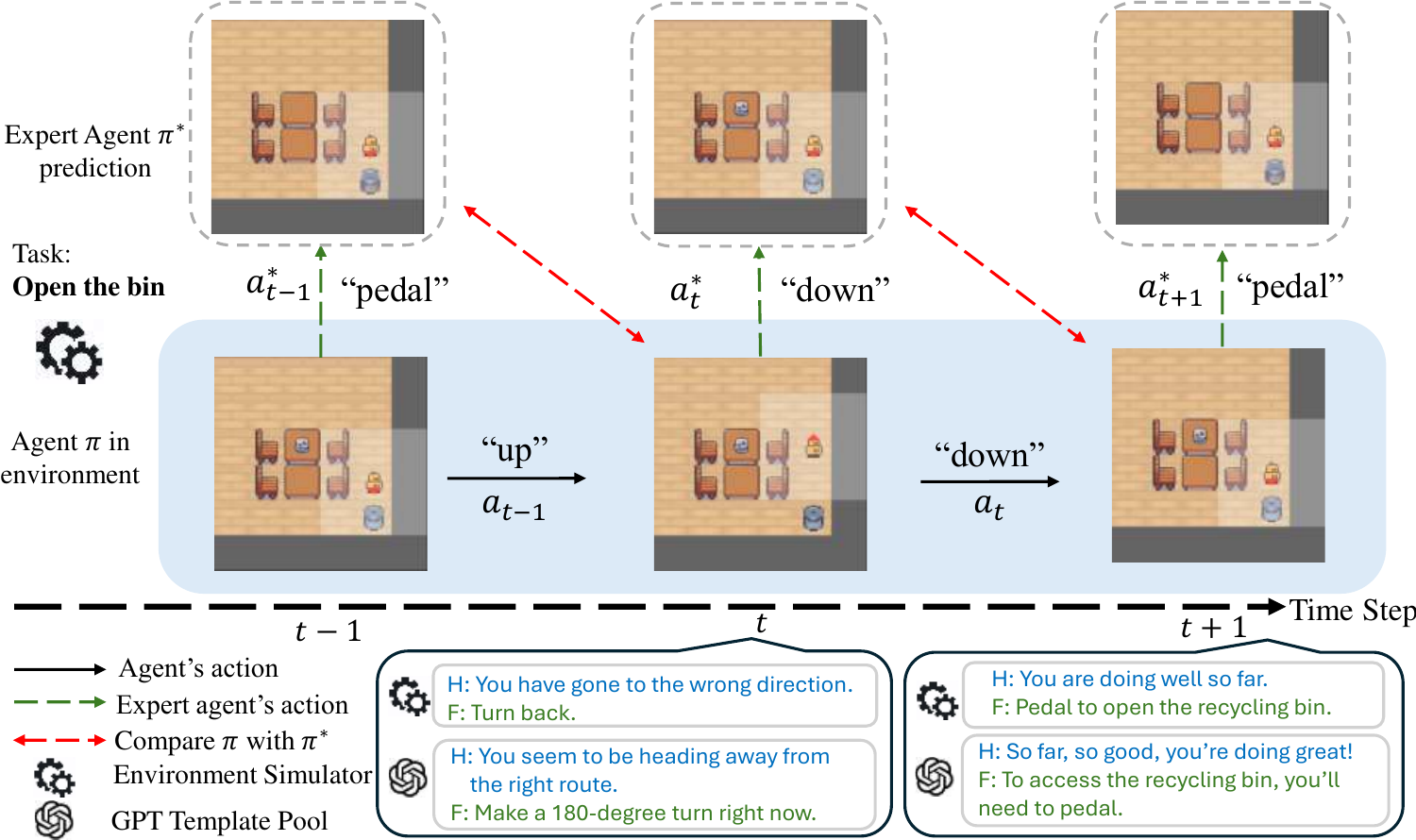}
    \vspace{-6pt}
    \caption{A demonstration of hindsight and foresight language feedback generation. In our framework, the agent $\pi$ executes the trajectory, while the expert agent $\pi^{*}$, with access to privileged ground truth knowledge, is used solely to provide information for generating language feedback to $\pi$. At time step $t$, \textcolor[RGB]{0, 112, 192}{hindsight language} is generated by comparing the agent's action $a_{t-1}$ with the expert agent's action $a_{t-1}^*$, whereas \textcolor{deepgreen}{foresight language} is generated by referring to the expert agent's action $a_{t}^*$ to guide the agent on the next step. To increase the diversity of language feedback, we construct a pool of language templates comprising GPT-augmented languages, and sample candidate instructions as online language feedback. 
    }
    \vspace{-15pt}
    \label{fig:hf_generation}
\end{figure*}
\begin{algorithm}[t]
\small
\caption{Offline Data Collection}
\label{alg:data_collection}
\begin{algorithmic}[1]
\State Initialize $\mathcal{D} \leftarrow \emptyset$

\For{each episode with $seed_i$}
    \State Initialize $\mathcal{D}_i \leftarrow \emptyset$
    \State Initialize environment $env$ with $seed_i$.
    \State Append task description $Td$ to $\mathcal{D}_i$
    \State Initialize the non-expert agent with a sub-optimal policy $\pi$.
    \State Initialize the expert agent with policy $\pi^{*}$.
    
    \For{each time step}
        \State $a_t \leftarrow \pi(h_{t})$
        \State $a^{*}_t \leftarrow \pi^{*}(h_{t})$
        \State $r_t, s_t, l_t^{hind}, l_t^{fore} \leftarrow \text{env}(a_t, a^{*}_t| h_t )$.
        \If{Use GPT-augmented Pool}
            \State $l_t^{hind}=\text{GPT-augmented}(l_t^{hind})$
            \State $l_t^{fore}=\text{GPT-augmented}(l_t^{fore})$
        \EndIf
        \State $l_t \gets \left\{
        \begin{array}{ll}
        l_t^{\text{hind}} + l_t^{\text{fore}} & \text{if H + F} \\
        l_t^{\text{hind}} & \text{if only H} \\
        l_t^{\text{fore}} & \text{if only F} \\
        \text{<empty>} & \text{if No Lang}
        \end{array}
        \right.$
        \State Append $(r_t, s_t, a_t, l_t)$ to $\mathcal{D}_i$
    \EndFor
    \State Aggregate Datasets $\mathcal{D} \leftarrow \mathcal{D}\cup\mathcal{D}_i$
\EndFor
\end{algorithmic}
\end{algorithm}

\subsection{Trajectory Generation}
To improve model generalization and avoid overfitting, it is essential to train on diverse, sub-optimal trajectories rather than relying solely on optimal ones generated by an expert agent \cite{kumar2020conservative,chen2021decision}. We achieve this by introducing perturbations to an expert planner (see Appendix \ref{Appendix:Agent_for_collection}), allowing the non-expert agent to produce sub-optimal trajectories. This promotes broader exploration of the state-action space, enhancing the model’s ability to generalize to unseen scenarios \cite{kumar2020conservative,chen2021decision}.

During data collection, we begin by appending the task description $Td$ to the trajectory sequence and initializing the environment with a fixed seed. A non-expert agent, using a sub-optimal policy $\pi$ derived from the expert agent’s optimal policy $\pi^{*}$, interacts with the environment. At each time step, the environment state $o_t$, reward $\hat{R}_t$, and the non-expert agent’s action $a_t$ are recorded to form the trajectory sequence: $(Td, \hat{R}_1, s_1, a_1, \ldots, \hat{R}_t, s_t, a_t)$.

\vspace{-5pt}
\subsection{Language Feedback Generation}
\vspace{-5pt}
For the second part of the dataset $\mathcal{D}$, we collect the language feedback along the non-expert agent's trajectory. As shown in Figure \ref{fig:hf_generation}, we follow a structured process to generate diverse and informative language feedback. For the state at time step $t$, the expert agent $\pi^*$ proposes an expert action $a^*_t$ (e.g. "down") at this state, which is further transformed into a foresight template \textcolor{deepgreen}{$l_t^{fore}$} (e.g. \textcolor{deepgreen}{"Turn back."}) by the environment simulator, guiding the agent on what should be done at this state. After the non-expert agent $\pi$ steps the environment (into time step $t+1$) with its generated action $a_t$ (e.g. "down"), the environment simulator generates a hindsight template \textcolor[RGB]{0, 112, 192}{$l_{t+1}^{hind}$} (e.g. \textcolor[RGB]{0, 112, 192}{"You are doing well so far."}) based on the comparison between agent action $a_t$ and expert agent action $a_t^*$ at the last time step $t$, reflecting on whether the agent is on the right track. 

For each foresight/hindsight template, we use GPT-4 to augment it into more natural and varied expressions. (e.g. We can augment "You are doing well so far." into \textit{"Up until now, you're doing wonderfully."} or \textit{"So far, so good, you're doing great!".}) We compile all the rewritten sentences into a set called the \textbf{\em GPT-augmented language pool}. At each step of the non-expert agent, we randomly select one candidate from the pool as the language instruction. This process ensures the feedback provided to the agent has high level of diversity and enriches the learning experience.

The level of informativeness and diversity of the language feedback depends on the inclusion of hindsight and foresight (e.g. concatenated when both are required) and the use of GPT-augmented language pool. The language feedback at each time step will finally get concatenated with the trajectory sequence into $(Td, \hat{R}_1, s_1, a_1, l_1, \ldots \hat{R}_t, s_t, a_t, l_t)$.  Algorithm \ref{alg:data_collection} summarizes the data collection process.

\vspace{-5pt}
\section{Model}
\vspace{-5pt}

\begin{figure}[ht]
\centering
\includegraphics[width=0.48\textwidth]{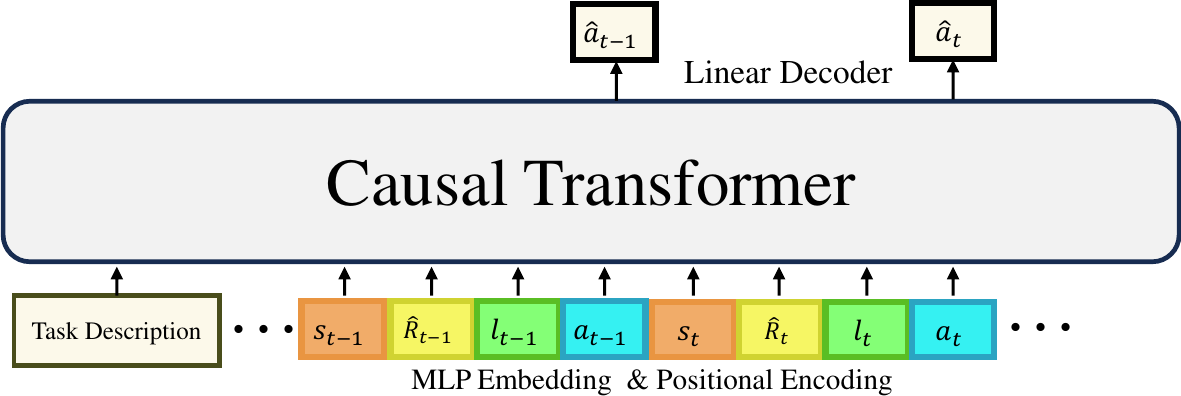}
\vspace{-15pt}
\caption{Language-Teachable Decision Transformer.}
\label{dt}
\vspace{-15pt}
\end{figure}
\noindent\textbf{Architecture.} We extend the Decision Transformer (DT) architecture \cite{chen2021decision} to create the Language-Teachable Decision Transformer (LTDT) by augmenting the input to include language feedback. This architecture is a decoder-only transformer, similar to GPT-2 \cite{radford2019language}, and models a trajectory sequence $(Td, \hat{R}_1, s_1, a_1, l_1, \ldots, \hat{R}_t, s_t, a_t, l_t)$, with the language feedback input appended at each step and a task description (TD) input prefixed at the beginning of the sequence. Like the original DT, the embeddings of these inputs are passed through the Causal Transformer, which encodes positional information to maintain sequence order. The transformer's output is used to predict the next action in the sequence, conditioned on the state, return-to-go, action, and language feedback in the last $K$ time steps, with the task description as the prefix ($4K+1$ tokens in total), as shown in Figure \ref{dt}.

\noindent\textbf{Training.} \ 
Similar to the original DT training, given an offline dataset of trajectory sequences, we sample a sub-sequence of length $K$ (with $4K+1$ tokens), and the prediction head is trained to predict discrete actions with the cross-entropy loss or continuous actions with the MSE loss. More training details can be found in Appendix \ref{models_and_training}.

\noindent\textbf{Language Embeddings.}
We use language embeddings from a frozen Sentence-BERT model \cite{reimers-2019-sentence-bert} in all environments. We find Sentence-BERT more sensitive to language feedback changes, capturing nuanced semantic differences better. 

\vspace{-5pt}
\section{Experiment}
\vspace{-5pt}

In this section, we design experiments to answer the following two research questions (RQs):
\vspace{-8pt}
\begin{itemize}
    \item \textbf{RQ 1}: How do the \textbf{\em informativeness} and \textbf{\em diversity} of language affect agents' performance on seen tasks?
    \item \textbf{RQ 2}: How does the \textbf{\em informativeness} of the language feedback affect pre-trained agents' adaptability on \textbf{\em unseen} tasks?
\end{itemize}
\vspace{-8pt}

For RQ1, we control agents trained with hindsight information, foresight information, or both to investigate the function of informativeness. We compare agents trained with language from both hand-crafted templates and the GPT-augmented language pool to examine the function of language diversity.

For RQ2, agents are taught in languages from the GPT-augmented language pool and tested on unseen tasks after fine-tuning with few-shot samples.

\begin{figure*}[ht]
\centering
\includegraphics[width=0.95\textwidth]{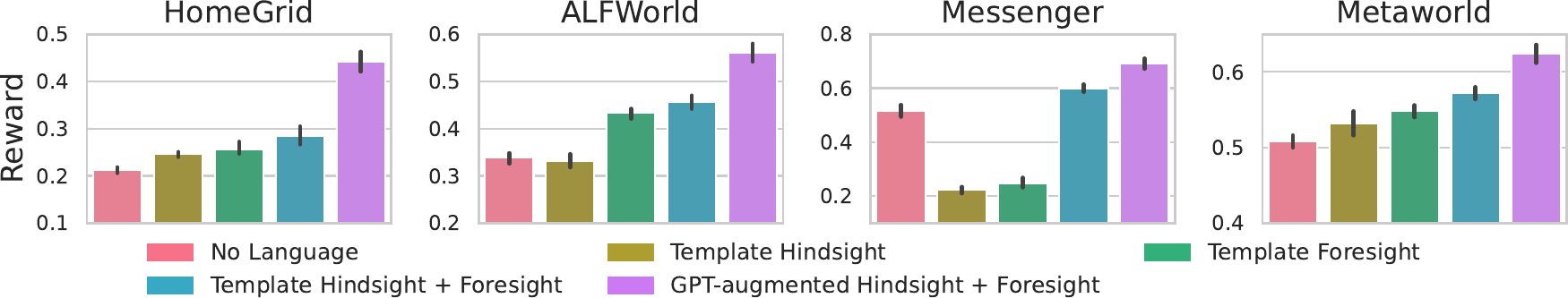}
\vspace{-8pt}
\caption{Comparison of agent performance in four environments (averaged across 100 seeds in each environment) under varying levels of language feedback informativeness and diversity. Agents trained with more informative language feedback exhibit progressively higher performance. Furthermore, given the same informativeness (Hindsight + Foresight), increasing diversity with the GPT-augmented language pool leads to the highest performance.}
\vspace{-8pt}
\label{hypo1}
\end{figure*}

\begin{figure*}[ht]
\centering
\includegraphics[width=1.0\textwidth]{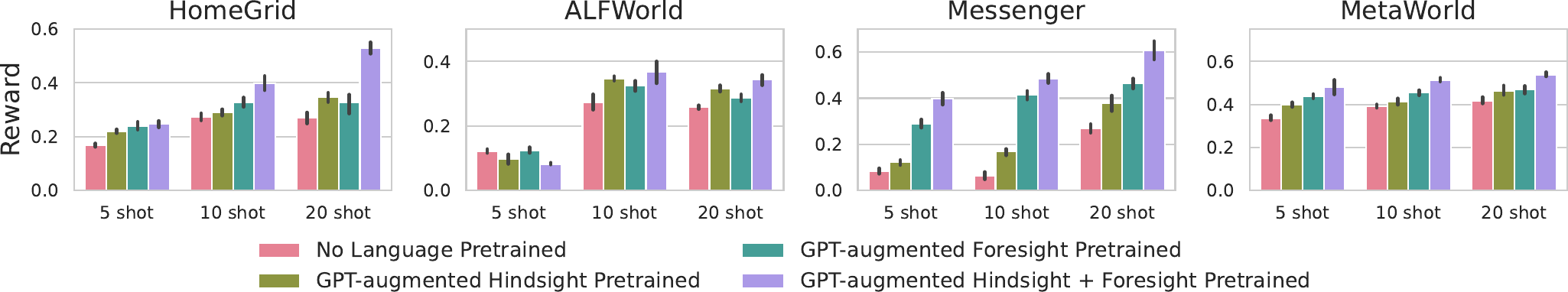}
\vspace{-20pt}
\caption{Comparison of agent performance on \textbf{unseen tasks} in four environments (averaged across 100 seeds in each environment) under varying language informativeness in agent pre-training. Agent trained with more informative language adapts to new tasks faster and better.}
\vspace{-18pt}
\label{fig:hypo2}
\end{figure*}

\subsection{Experimental Setup}

\noindent\textbf{Setup for RQ 1.} We compare performance on seen tasks between agents trained with varying levels of language informativeness and diversity: 1) the \texttt{No Language} agent is trained without any language instructions; 2) the \texttt{Template Foresight} agent is trained with hand-crafted foresight language templates; 3) the \texttt{Template Hindsight} agent is trained with hand-crafted hindsight language templates; 4) the \texttt{Template Hindsight + Foresight} agent is trained with hand-crafted foresight and hindsight language templates; and 5) the \texttt{GPT-augmented Hindsight + Foresight} agent is trained with hindsight and foresight languages from the GPT-augmented language pool. 
We train on 100, 1,000, 20,000, and 10,000 trajectories for HomeGrid, ALFWorld, Messenger, and MetaWorld environments, respectively. Evaluation is performed over 5 runs, with 100 random seeds for each run.

\noindent\textbf{Setup for RQ 2.} We pre-train different agents on seen tasks and then compare adaptability (how well an agent performs after few-shot learning) on unseen tasks: 1) the \texttt{No Language pre-trained} agent is pre-trained without any language instructions; 2) the \texttt{GPT-augmented hindsight pre-trained} agent is pre-trained with hindsight language from the GPT-augmented language pool; 3) the \texttt{GPT-augmented foresight pre-trained} agent is pre-trained with foresight language from the GPT-augmented language pool; 4) the \texttt{GPT-augmented hindsight + foresight pre-trained} agent is pre-trained with both hindsight and foresight language from the GPT-augmented language pool.  During the few-shot adaptation stage, we choose to fine-tune the pre-trained agents with both hindsight + foresight language from the GPT-augmented language pool for all settings, since this mimics a real-world few-shot learning scenario, where humans likely provide diverse feedback, including both hindsight and foresight, to guide the agent in new tasks. 
We pretrain on 6,432, 1,000, 20,000, and 10,000 trajectories for HomeGrid, ALFWorld, Messenger, and MetaWorld, respectively. For all environments, we adapt on 5, 10, and 20 trajectories to 1 new task. Evaluation is performed over 5 runs, with 100 seeds per run.

Further details on task setup of RQ 1 and RQ 2 can be found in Appendix \ref{sec:task_setting}. Additional results when training and adapting on same types of language can be found in Appendix \ref{align_appendix}.

\noindent\textbf{Evaluation.}
\label{sec:eval_lang_feedback}
At inference time, an agent is given a short task description before it starts to act, and language feedback along its execution. The language feedback should ideally come from real humans, who provide feedback varying in informativeness, diversity, and frequency (how often feedback is provided). However, recruiting and moderating real humans to generate online feedback is expensive and difficult to scale. Therefore, we employ GPT-4 to provide online language feedback to mimic real humans. Specifically, at each time step, we provide all necessary context information to GPT-4 in its prompt and let it decide ``whether to speak'' (frequency), ``what to speak'' (informativeness), and ``how to speak'' (diversity). The context information, in this case, consists of the ground-truth environment states, action/state history, and template-based hindsight and foresight short text description generated by comparing the actions of the expert agent and the trained agent. GPT-4 then has the freedom to rephrase, combine, shorten, and discard such context information to utter diverse, coherent, and natural language feedback, mimicking a real human. See Appendix \ref{Appendix:Language Feedback in Evaluation} for an example of such GPT-generated online feedback.

\noindent\textbf{Metric.} We use the reward value as our main metric. Agents receive a reward of 1 upon task completion for all environments and receive additional rewards for achieving specific sub-goals for the HomeGrid and ALFWorld environments.

\subsection{Experimental Results}

\label{main_result}
\begin{figure*}[tp]
\centering
\includegraphics[width=\linewidth]{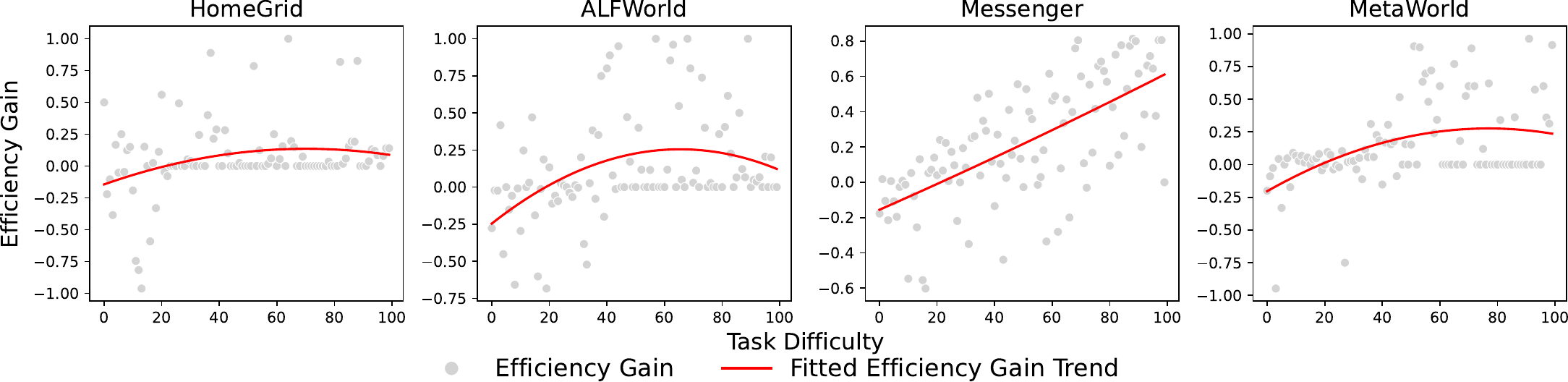}
\vspace{-22pt}
\caption{Efficiency gain vs. task difficulty. We fit the scatter plots with a second-degree polynomial to visualize the overall trend. As task difficulty increases, the general trend of the efficiency gain is to rise initially and then decline, suggesting: (1) for tasks that are too easy or too hard, language feedback does not improve efficiency; (2) language feedback is most helpful in increasing efficiency for moderate tasks.}
\label{fig:ablation1}
\vspace{-18pt}
\end{figure*}

\noindent\textbf{Results for RQ 1.} As we can see in Figure \ref{hypo1}, agents trained with both diverse and informative language feedback (\texttt{GPT-augmented Hindsight + Foresight}) consistently achieve the highest performance across all environments. The varied and paraphrased instructions generated from GPT provide a richer set of linguistic inputs, enabling the agents to develop a more robust language understanding for task execution during evaluation.

When examining the impact of informativeness, we observe that agents trained with both hindsight and foresight information (Template Hindsight + Foresight) consistently achieve higher performance across all environments compared to those trained with only hindsight or foresight information. This indicates that integrating both types of feedback enhances the informativeness of the language, enabling the agents to develop a more comprehensive understanding and leading to better decision-making and overall performance. The only exception is in the Messenger environment, where the no-language agent exhibits a surprisingly strong performance. However, upon further investigation of this exception, we find that if the hindsight-only or foresight-only feedback is from the GPT-augmented pool, the agent can still outperform the \texttt{No Language} agent  (refer to Appendix \ref{sec:messager-res}).

In terms of diversity, the results show that agents trained with diverse language feedback, as indicated by the `GPT-augmented' bars, consistently outperform those trained with less varied language input. The rich set of augmented instructions generated by GPT helps agents develop a more flexible and nuanced understanding of task instructions, which translates to better performance during evaluation. This highlights the critical role of linguistic diversity in enhancing the robustness and adaptability of the agents' language comprehension, ultimately leading to improved task execution across different environments.

\begin{figure}[tp]
\centering
\includegraphics[width=0.87\linewidth]{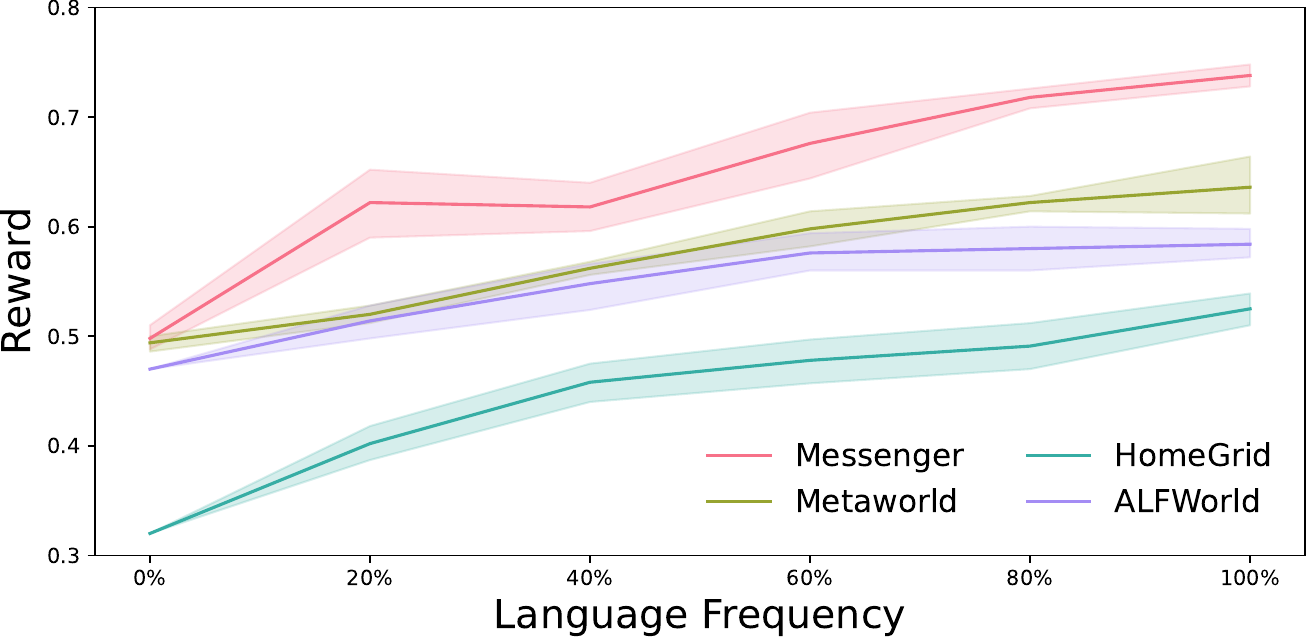}
\vspace{-8pt}
\caption{Performance vs. language frequency. Agents perform better with more frequent language feedback across four environments.}
\vspace{-14pt}
\label{fig:ablation2}
\end{figure}

\noindent\textbf{Results for RQ 2.} 
The results in Figure \ref{fig:hypo2} reveal that agents pre-trained with more informative language can adapt to unseen tasks \emph{faster} and \emph{better}. 
``Adapting faster'' is evident by the fact that agents pre-trained with GPT-augmented Hindsight + Foresight language in 5 or 10 shots can already achieve a similar performance 20-shot performance of agents trained with less informative language. ``Adapting better'' is evident by the fact that, at a given number
of shots available for adaptation, the agent trained
with the most informative language performs the
best compared to its less informatively-pretrained
counterparts.
These results indicate that agents pre-trained with more informative language can adapt and generalize to new tasks faster and better.

\vspace{-8pt}
\subsection{Ablation Study}
\vspace{-5pt}
\paragraph{Efficiency Gain vs. Task Difficulty.} Can language feedback help the agent to achieve more different tasks? To answer this question, we define \textbf{efficiency gain} as the difference in efficiency between an agent trained with informative and diverse GPT languages, and an agent trained without any languages. Efficiency is measured by a path-weighted reward, as introduced in ALFRED \cite{shridhar2020alfred}. This reward, $r_p$, is calculated as $r_p = r \times \frac{L^*}{\max(L, L^*)}$, where $r$ is the total reward, $L$ is the agent's trajectory length, and $L^*$ is the expert agent's trajectory length. Higher $r_p$ indicates successful task completion with fewer steps. 

We define \textbf{task difficulty} for each configuration by calculating the average success rates of agents trained without language feedback, ranking these from lowest to highest. Configurations with lower success rates are considered more difficult, indicating greater challenges for agents learning from these configurations without language assistance.
As shown in Figure \ref{fig:ablation1}, the efficiency gain generally rises with increasing learning difficulty, then declines. This suggests that: (1) for tasks that are too easy or too hard, language feedback does not improve efficiency; (2) language feedback is most helpful in increasing efficiency for moderate tasks.

\noindent\textbf{Performance vs. Language Frequency.}
In the main experiments, we utilize an online GPT model to determine whether to provide language feedback at each time step. However, it is important to explore how varying the frequency of language feedback influences agent performance. To investigate this, we control the feedback frequency by sampling according to pre-defined probabilities (e.g., 20\%, 40\%). The language feedback is extracted from the GPT-augmented language pool; if no language is sampled, an empty string is provided to the agent. The evaluation is conducted on agents trained with both hindsight and foresight feedback derived from the GPT-augmented language pool.
As illustrated in Figure \ref{fig:ablation2}, agents' performance improves steadily across all environments with more frequent language feedback during evaluation. This finding suggests that agents trained with informative and diverse language feedback can continually absorb and leverage new information when additional feedback is provided, leading to enhanced performance.

\begin{figure}
\centering
\includegraphics[width=\linewidth]
{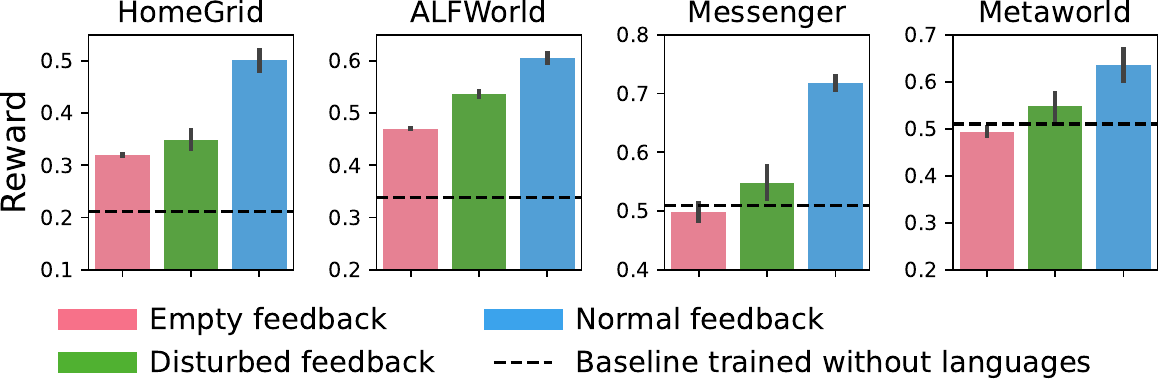}
\vspace{-20pt}
\caption{We investigate two special evaluation settings: (1) no language feedback is provided during evaluation and (2) disturbed language feedback is given at every step. Results show that agents trained with the GPT-augmented language still outperform the no-language agent (the black dotted line) in the disturbed setting, and also achieve better performance in some environments while no language is given.
\vspace{-15pt}
}
\label{fig:ablation3}
\end{figure}

\noindent\textbf{Performance under Corrupted Language.} This ablation aims to evaluate how agents perform when provided with incorrect instructions. We assess the performance of an agent trained with GPT-4-augmented informative and diverse language under two conditions: (1) Empty Feedback: the absence of language feedback during testing, and (2) Disturbed Feedback: the provision of disturbed language at each step. The disturbed language consists of redundant, irrelevant, or misleading information (e.g., incorrect actions or objects) and is generated using GPT-augmented templates with disrupted content.
The results in Figure \ref{fig:ablation3} reveal two interesting findings:
(1) When tested without any language feedback, the agent trained with informative and diverse language performs comparably or even exceeds the performance of the agent trained without any language (represented by the black dotted line). This indicates that the agent develops a robust intrinsic understanding of the task, demonstrating that it does not overly rely on language feedback;
(2) When exposed to disturbed feedback, the agent trained with informative and diverse language maintains performance levels comparable to the no-language agent. This showcases the agent’s ability to withstand misleading information, a critical trait for real-world applications where human feedback may be unreliable.

\vspace{-5pt}
\section{Conclusion}
In this paper, we investigate how the informativeness and diversity of language feedback affect embodied agents. We introduce the Language-Teachable Decision Transformer (LTDT), which makes decisions based on human language feedback. To facilitate the training of LTDT agents, we propose an easy-to-use pipeline for collecting offline hindsight and foresight GPT templates. We compare the performance of agents by varying the informativeness and diversity of the training languages across four reinforcement learning environments and evaluate the agents' ability to understand real-world human language using online GPT as a proxy. Our results demonstrate that training with more informative and diverse language feedback significantly enhances agent performance and enables fast adaptation to unseen tasks.

\section*{Limitations}
Our study has several limitations. First, the investigated environments are primarily game-based and do not test the agents' ability to incorporate real-life visual inputs. Future work will focus on evaluating agents in more realistic and complex environments that involve real-world visual inputs and challenges. Second, while GPT language outputs can produce diverse and contextually relevant language, they may not fully cover all human language styles and nuances. Specifically, GPT models might miss certain idioms, dialects, or culturally specific references that are prevalent in human communication. Future work will aim to incorporate a broader spectrum of language variations and test agents in scenarios involving more diverse linguistic inputs.

\vspace{-5pt}
\section*{Ethical Impacts}
\vspace{-5pt}
Our study, conducted entirely within simulated environments, does not present immediate ethical concerns. The teachable nature of our Language-Teachable Decision Transformer (LTDT) method is designed to make AI agents more controllable and better aligned with human values, promoting safer and more ethical interactions. By enhancing agent performance through informative and diverse language instructions, we aim to foster AI systems that are more transparent and responsive to human guidance, addressing ethical considerations in the deployment of artificial intelligence. As AI becomes more mainstream, these considerations are increasingly pertinent, and our work strives to advance AI technology responsibly.

\vspace{-5pt}
\section*{Acknowledgements}
\vspace{-5pt}
This work was supported by NSF IIS-1949634 and has benefited from the Microsoft Accelerate Foundation Models Research (AFMR) grant program.
We would like to thank the anonymous reviewers for their valuable comments and suggestions.

\bibliography{ref}
\appendix
\clearpage
\section{Environments and Language Feedback}
\label{Appendix:Language_feedback}
\subsection{Environments Overview}
The Appendix Table \ref{tab:env_info} lists the information that is inherently available within the environment. All models, regardless of whether they are trained with language input or not, will have access to this environmental information. 
\begin{table}[ht]
    \resizebox{0.5\textwidth}{!}{%
    \begin{tabular}{l c c c}
        \hline
        \textbf{Env} & \textbf{Image Observation} & \textbf{Instruction Manual} & \textbf{Text State Description} \\ \hline
        HomeGrid   & Yes  & No  & No  \\
        AlfWorld   & No   & No  & Yes \\
        Messenger  & No   & Yes  & No \\
        MetaWorld  & No   & No  & No \\ \hline
    \end{tabular}
    }
    \caption{Information provided by each environment.}
    \label{tab:env_info}
    \vspace{-12pt}
\end{table}

\subsection{Language Feedback for Different Environments}
For each environment, we design multiple templates conveying different meanings, and then applied GPT-4 to augment the languages into a GPT-augmented language pool. The number of templates and the corresponding GPT-augmented sentences for each template are shown in Appendix Table \ref{tab:template}.
\begin{table}[ht]
    \resizebox{0.5\textwidth}{!}{%
    \begin{tabular}{l c c c}
        \hline
        \textbf{Env} & \textbf{\# Hind Templates} & \textbf{\# Fore Templates} & \textbf{\# AUG} \\ \hline
        HomeGrid   & 20  & 9  & 70  \\
        AlfWorld   & 4   & 4  & 200 \\
        Messenger  & 4   & 4  & 80 \\
        MetaWorld  & 2   & 6  & 180 \\ \hline
    \end{tabular}
    }
    \caption{Number of templates and augmented sentences for each environment, where '\# Hind Templates' refers to the number of hindsight templates, '\# Fore Templates' refers to the number of foresight templates, and '\# AUG' refers to the number of GPT-augmented sentences per template.}
    \label{tab:template}
    \vspace{-12pt}
\end{table}

\subsubsection{HomeGrid}
\label{HomeGrid_language}
HomeGrid is a multitask grid world designed to evaluate how well agents can understand and use various types of language to complete tasks. Agents will receive both task specifications and language hints, providing prior knowledge about world dynamics, information about world states, or corrections to assist the agents. We adopt the language hints in HomeGrid as foresight and further extend the environment to provide hindsight that provides comments on agents' past performance. Agents are expected to ground both hindsight and foresight to the environment to achieve higher performance. It includes five task types involving interaction with objects and bins (find, get, clean up, rearrange, open), with a total of 38 tasks. Object locations, bin locations, and bin dynamics are randomized. The agent receives a reward of 1 when the task is completed, and receives a reward of 0.5 if a subgoal exists (e.g., get the object in the clean-up task) and gets completed. Each template language is augmented to 70 sentences in the GPT template pool. Examples of hindsight and foresight languages are as follows:

\begin{enumerate}
    \item[\textbullet] Hindsight Examples: 
    \begin{itemize}
        \item[{\includegraphics[width=0.35cm]{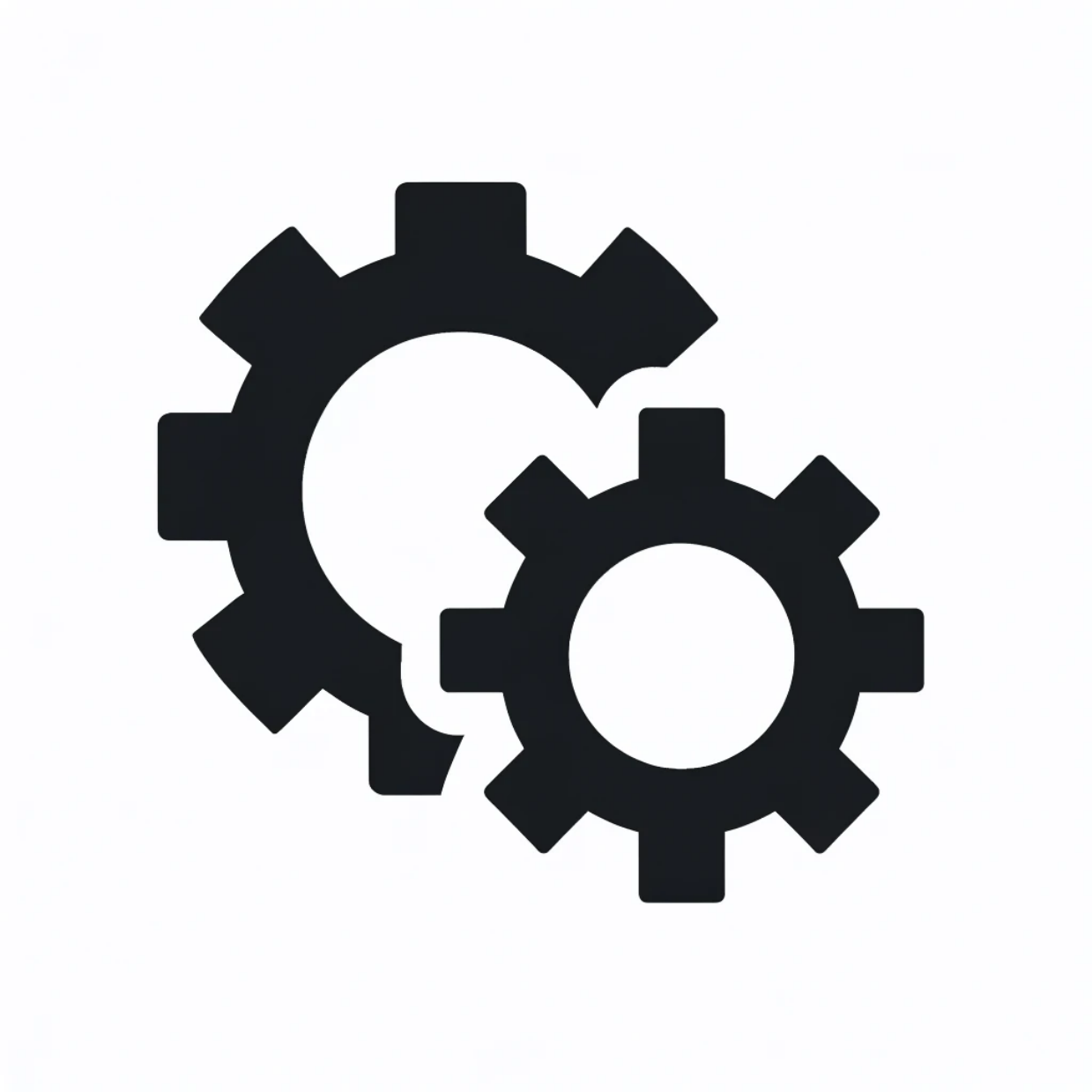}}] Template: 
        \begin{itemize}
            \item[\(\triangleright\)] "You have gone to the wrong direction."
            \item[\(\triangleright\)] "You are doing well so far."
        \end{itemize}
        \item[{\includegraphics[width=0.3cm]{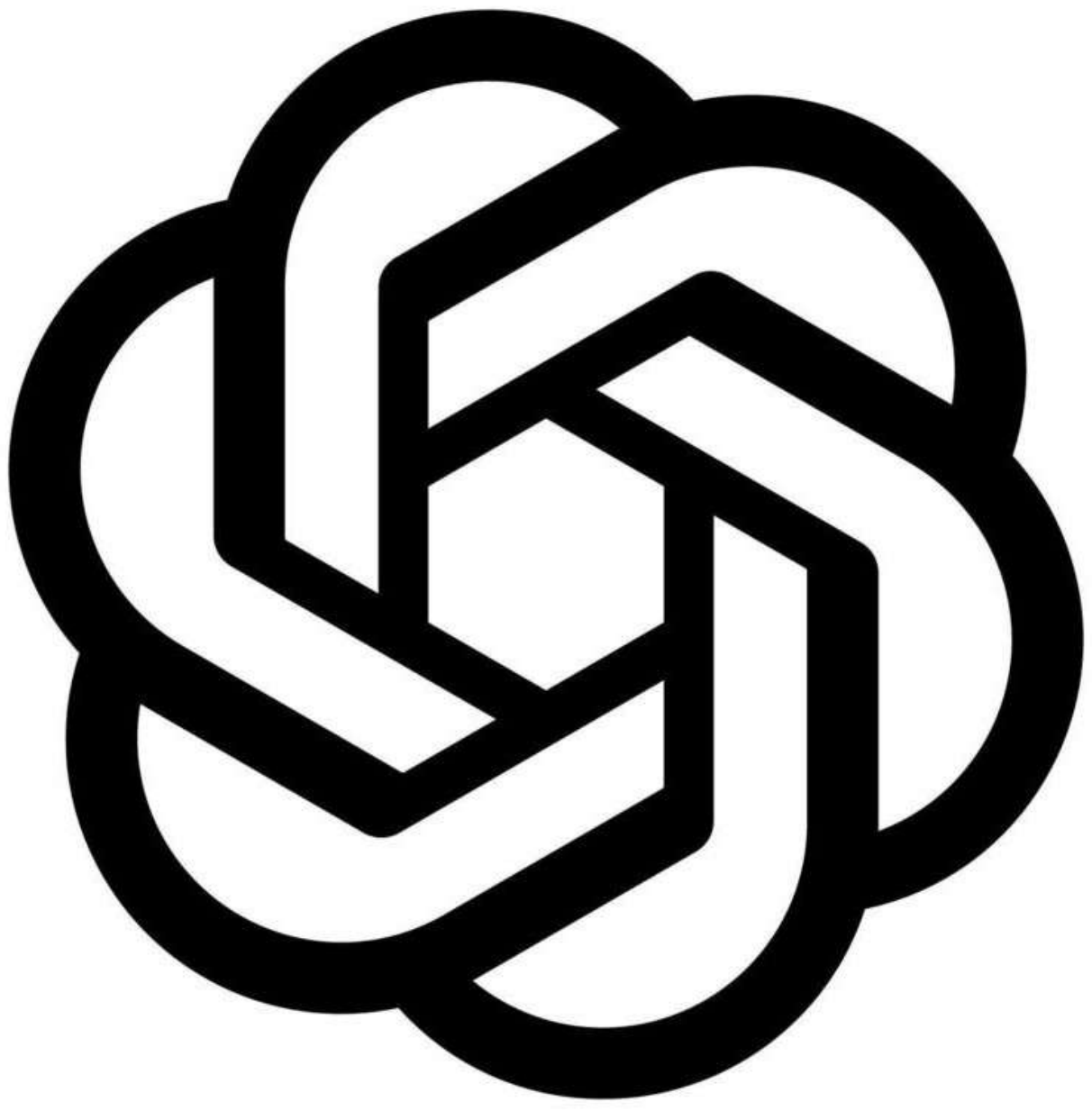}}]GPT Template: 
        \begin{itemize}
            \item[\(\triangleright\)] "You seem to be heading away from the right route."
            \item[\(\triangleright\)] "So far, so good, you are doing great!"
        \end{itemize}  
    \end{itemize}
    \item[\textbullet] Foresight Examples: 
    \begin{itemize}
        \item[{\includegraphics[width=0.35cm]{figures/env_simulator.pdf}}] Template: 
        \begin{itemize}
            \item[\(\triangleright\)] "Turn back."
            \item[\(\triangleright\)] "Pedal to open the recycling bin."
        \end{itemize}
        \item[{\includegraphics[width=0.3cm]{figures/gpt.pdf}}]GPT Template: 
        \begin{itemize}
            \item[\(\triangleright\)] "Make a 180-degree turn right now."
            \item[\(\triangleright\)] "To access the recycling bin, you'll need to pedal."
        \end{itemize}  
    \end{itemize}
\end{enumerate}
Language instructions are generated based on the comparison of agent's action and expert planer action, considering distance, relative location, and interaction between the agent and target objects.

\subsubsection{ALFWorld}
\label{AlfWorld_language}

ALFWorld is a text-game environment that aligns with the embodied ALFRED benchmark \cite{shridhar2020alfred} and provides simulation for household tasks. It includes six types of tasks where agents need to navigate and interact with household objects through text actions. The location of the task objects is randomly located among 50 locations in each episode, making the task challenging for the agent to plan and for the subgoals.
For the experiment, we adopt LLF-ALFWorld \cite{cheng2023llf}, which provides an extra language wrapper for hindsight and foresight language generation over the original ALFWorld. The languages are generated based on both agents' past actions and the optimal trajectory for the current episode. Agent gets a reward of 1 when the task is completed. Each template is augmented to 200 sentences in GPT template pool. Examples of hindsight and foresight languages are as follows:

\begin{enumerate}
    \item[\textbullet] Hindsight Examples: 
    \begin{itemize}
        \item[{\includegraphics[width=0.35cm]{figures/env_simulator.pdf}}] Template: 
        \begin{itemize}
            \item[\(\triangleright\)] "You made a mistake by taking the bad action \{action\}."
            \item[\(\triangleright\)] "It was a right decision to not take the bad action \{action\}."

        \end{itemize}
        \item[{\includegraphics[width=0.3cm]{figures/gpt.pdf}}]GPT Template: 
        \begin{itemize}
            \item[\(\triangleright\)] "The choice to implement \{action\} was misguided."
            \item[\(\triangleright\)] "You made a sensible choice by not committing to the \{avoid action\}."
        \end{itemize}  
    \end{itemize}
    \item[\textbullet] Foresight Examples: 
    \begin{itemize}
        \item[{\includegraphics[width=0.35cm]{figures/env_simulator.pdf}}] Template: 
        \begin{itemize}
            \item[\(\triangleright\)] "You should now take the \{action\} action."
            \item[\(\triangleright\)] "Take \{action\} in the next step."
        \end{itemize}
        \item[{\includegraphics[width=0.3cm]{figures/gpt.pdf}}]GPT Template: 
        \begin{itemize}
            \item[\(\triangleright\)] "Consider taking the \{action\} as your next step."
            \item[\(\triangleright\)] "Moving on, consider the \{action\} action."
        \end{itemize}  
    \end{itemize}
\end{enumerate}
Language instructions are generated based on experts' next action and whether agent's past actions are aligned with expert past actions, considering whether agents have moved to the target position and conducted correct interaction with the objects.

\subsubsection{Messenger}
\label{Messenger_language}
Messenger is a grid world with several entities. The agent's primary task is to retrieve a message from one entity and deliver it to another goal entity, all while avoiding enemies. At the start of each episode, the agent is provided with a manual describing the randomized roles of the entities and their movement dynamics.  The challenge lies in the fact that the agent does not have access to the true identity of each entity and must ground the text manual to the dynamics, necessitating multi-hop reasoning. (For example, grounding the "an approaching queen is a deadly enemy" to the observations of dynamics.) \cite{lin2023learning} The agent receives a sparse reward of 1 when the task is completed. Each template language is augmented to 80 sentences in the GPT template pool. Examples of hindsight and foresight languages are as follows:
\begin{enumerate}
    \item[\textbullet] Hindsight Examples: 
    \begin{itemize}
        \item[{\includegraphics[width=0.4cm]{figures/env_simulator.pdf}}] Template: 
        \begin{itemize}
            \item[\(\triangleright\)] "It's good that you are getting close to the \{target\} at \{target direction\} by moving \{direction\}!"
            \item[\(\triangleright\)] "Stepping \{action direction\}, yet you ran into \{enemy name\}. Be more cautious."

        \end{itemize}
        \item[{\includegraphics[width=0.3cm]{figures/gpt.pdf}}]GPT Template: 
        \begin{itemize}
            \item[\(\triangleright\)] "Good job on approaching the \{target\} to the \{target direction\} by moving \{direction\}! "
            \item[\(\triangleright\)] "Stepping \{action direction\} directly met \{enemy name\}. Needs strategic thinking." 
        \end{itemize}  
    \end{itemize}
    \item[\textbullet] Foresight Examples: 
    \begin{itemize}
        \item[{\includegraphics[width=0.4cm]{figures/env_simulator.pdf}}] Template: 
        \begin{itemize}
            \item[\(\triangleright\)] "Move \{optimal direction\} to approach the \{target name\} located at the \{target direction\}. "
            \item[\(\triangleright\)] "Rest assured, there are no enemies around."
        \end{itemize}
        \item[{\includegraphics[width=0.3cm]{figures/gpt.pdf}}]GPT Template: 
        \begin{itemize}
            \item[\(\triangleright\)] "To get to the \{target name\} at \{target direction\}, go \{optimal direction\}. "
            \item[\(\triangleright\)] "Not detecting any danger, it's safe."
        \end{itemize}  
    \end{itemize}
\end{enumerate}
When generating the language instructions, we compare the agent's actions and the expert's actions, considering the locations of the target and nearest enemy, calculating the distance and generate the hindsight reflections based on some engineered rules.

\subsubsection{MetaWorld}
\label{MetaWorld_language}
MetaWorld is a simulated benchmark that includes a variety of manipulation tasks performed using a Sawyer robot arm. It includes 50 types of robot manipulation tasks common in daily life. Since our main goal is not meta-learning, we select the "assembly" and "hammer" tasks for pretraining and adaptation in our experiments. This requires the agent to pick up the tool and aim at the specific target with high precision. To increase the challenge of the tasks, we introduce random disturbances at random steps. This requires the robot to actively recover and return to its normal trajectory whenever it deviates. The agent receives a sparse reward of 1 when completing the task. Each template language is augmented to 180 template languages in the GPT template pool. Examples of hindsight and foresight languages are shown in the following:
\begin{enumerate}
    \item[\textbullet] Hindsight Examples: 
    \begin{itemize}
        \item[{\includegraphics[width=0.35cm]{figures/env_simulator.pdf}}] Template: 
        \begin{itemize}
            \item[\(\triangleright\)] "It's excellent to raise the gripper."
            \item[\(\triangleright\)] "You are making mistakes for not opening your gripper."

        \end{itemize}
        \item[{\includegraphics[width=0.3cm]{figures/gpt.pdf}}]GPT Template: 
        \begin{itemize}
            \item[\(\triangleright\)] "Good job for raising your gripper."
            \item[\(\triangleright\)] "You make a regrettable mistake since your gripper is closing."
        \end{itemize}  
    \end{itemize}
    \item[\textbullet] Foresight Examples: 
    \begin{itemize}
        \item[{\includegraphics[width=0.35cm]{figures/env_simulator.pdf}}] Template: 
        \begin{itemize}
            \item[\(\triangleright\)] "It's time to grasp the wrench now."
            \item[\(\triangleright\)] "Please raise the hammer."
        \end{itemize}
        \item[{\includegraphics[width=0.3cm]{figures/gpt.pdf}}]GPT Template: 
        \begin{itemize}
            \item[\(\triangleright\)] "Can you grab the wrench with your gripper?"
            \item[\(\triangleright\)] "I think the hammer should be raised now."
        \end{itemize}  
    \end{itemize}
\end{enumerate}
We compare the agent's actions with the expert's actions, and tell the agent's whether their decisions at the previous step matches with the expert's actions, and inform them of what an expert will do at the next step.

\section{Agent for Offline Data Collection and Language Feedback Generation}
\label{Appendix:Agent_for_collection}
 We use an expert agent and a non-expert agent with sub-optimal policies during the data collection. The sub-optimal policy is used for introducing some errors or perturbations in the training data, and letting the expert policy continue to recover. This helps agents learn to recover from potential failures using hindsight reflections and foresight instructions. In our experiments, we introduced 10-20\% random noise in each trajectory as a sub-optimal policy. We found that this level of perturbation aids learning, but excessive disturbance (e.g., >50\% per trajectory) significantly degrades performance as agents start learning suboptimal behaviors.
\subsection{HomeGrid}
For the HomeGrid environment, we design an expert planer to work as the expert agent. We first divide the task into several sub-tasks (i.e. divide "open the recycling bin" into 1. "navigate to the bin", 2. "open the bin"). For navigation (move to some place) sub-tasks, we implement breadth-first search to find the optimal path; for interaction sub-task (interact with object), we output the corresponding action. We implement the non-expert agent by adding "perturbation" into the expert planer. For example, we randomly reverse the next step of expert action and let the expert planner recover from the error. 

\subsection{ALFWorld}
For the ALFWorld environment, we use a pre-built expert planer from LLF-Bench \cite{cheng2023llf} to work as both the expert agent and the agent for the data collection. 

\subsection{Messenger}
As for the Messenger environment, we implement an expert agent using the A* algorithm \cite{Hart1968}. We define the cost by the distance to the target and the distance to the nearest enemies, and then heuristically search in the grid environment. The non-expert agent in the data collection is implemented by adding random disturbance to the expert agent.

\subsection{MetaWorld}
We build the expert agent on the pre-defined policy from the original MetaWorld codebase \cite{yu2019meta} and adapt the policy to random disturbance so that the expert planner can recover to a normal trajectory in any situation. 

\section{Task Settings for RQ 1 and 2}
\label{sec:task_setting}
\noindent\textbf{Task Setting for RQ 1.} We evaluate the agents' performance using the \emph{same tasks} as in the training phase (but with different initialization of the agents and object layout for different episodes). Concretely, 1) in HomeGrid, we train and evaluate on multi-tasks, including \textsc{find, get, rearrange} and \textsc{open}; 2) in ALFWorld, we train and evaluate on multi-tasks including \textsc{pick\&place, clean\&place} and \textsc{heat\&place} tasks; 3) in Messenger, we train and evaluate on the task goal \textit{``first retrieve the message and then deliver to target entity''}; and 4) in MetaWorld, we train and evaluate on the \textsc{assembly} task, in which the robot arm needs to pick up the wrench and put it on the peg. 

\noindent\textbf{Task Setting for RQ 2.} We evaluate agents' performance on \emph{unseen tasks} by first pre-training agents on certain tasks and then adapting agents to unseen tasks with few-shot episodes.  
Specifically, 1) in HomeGrid, we take \textsc{find, get, rearrange, open} tasks for pre-training and the \textsc{clean-up} task for adaptation and evaluation; 2) in ALFWorld, we take \textsc{pick\&place} and \textsc{clean\&place} for pretraining and \textsc{heat\&place} tasks for adaptation and evaluation; 3) in Messenger, we take \textit{``first retrieve the message and then deliver to target entity"} as the pretraining task and \textit{``first get to the target entity and then retrieve the message"} (where the order of the goal is reversed compared to the pretraining tasks) for adaptation and evaluation; 4) in MetaWorld, we take the \textsc{assembly} task for pretraining, and the \textsc{Hammer} task for adaptation and evaluation. 
\section{Performance under aligned language type with training.}\label{align_appendix} As stated in Section \ref{sec:eval_lang_feedback}, we use online GPT for all evaluations in RQ 1 and 2 to mimic real-life human language environments. In this section, we align the evaluation language type (and adaptation language type in RQ 2) with each agent's corresponding training language type for further investigation (e.g. \texttt{No Language} Agent is evaluated with empty language; \texttt{Template Hindsight} Agent is evaluated with Template Hindsight).
Experiments on RQ 1 and 2 are conducted on HomeGrid and Messenger respectively, with the results presented in Table \ref{tab:aligned_eval}. 
\begin{table}[h]
    \resizebox{0.48\textwidth}{!}{
    \begin{tabular}{l c c}
        \toprule[2pt]
        \multicolumn{3}{c}{\textbf{HomeGrid Env on RQ 1}} \\
        \textbf{Training Language} & \textbf{Aligned Eval} & \textbf{Online GPT Eval} \\ 
        \midrule[0.3pt]
        No Lang                                & 0.235  & 0.212 \\
        Template H                     & 0.260  & 0.246 \\
        Template F                     & 0.305  & 0.262 \\
        Template H + F         & 0.325  & 0.285 \\
        GPT-augmented H + F    & 0.472  & 0.442 \\ 
        \midrule[1pt]
        \multicolumn{3}{c}{\textbf{Messenger Env on RQ 2 (20 Shots)}} \\
        \textbf{Training Language} & \textbf{Aligned Adapt \& Eval} & \textbf{Online GPT Eval} \\ 
        \midrule[0.3pt]
        No Lang                                & 0.323  & 0.270 \\
        GPT-augmented H                & 0.450  & 0.378 \\
        GPT-augmented F                & 0.512  & 0.464 \\
        GPT-augmented H + F    & 0.623  & 0.608 \\ \bottomrule[2pt]
    \end{tabular}
    }
    \caption{Comparison of agents' performance adapted (for RQ 2) and evaluated with aligned language type in HomeGrid environment on RQ 1 and Messenger environment on RQ 2. `Aligned (Adapt \&) Eval' refers to (adaptation \&) evaluation with same type of language in training and `Online GPT Eval' refers to online GPT evaluation (results in Section \ref{main_result}). The results show that GPT-augmented Hindsight + Foresight evaluated with online GPT still outperforms other training settings even with aligned language evaluation, indicating higher language informativeness and diversity enhance intrinsic task understanding.}
    \label{tab:aligned_eval}
    \vspace{-12pt}
\end{table}

The results Table \ref{tab:aligned_eval} show that: (1) aligning the informativeness and diversity levels between training, adaptation and evaluation improves the final performance for all types; (2) more importantly, even with aligned evaluation and adaptation language, \textbf{no other settings have outperformed GPT-augmented Hindsight + Foresight evaluated with online GPT}. This further demonstrates that high informativeness and diversity in training language help agents intrinsically understand tasks to achieve better performance.
\section{Impact of hindsight on future steps}
\label{sec:hindsight-impact}
Compared to foresight feedback, which provides instructions for the correct action in the next step, hindsight feedback reflects on incorrect actions taken in previous steps. This retrospective analysis can still guide agents toward success by narrowing down the search space for corrective actions. To demonstrate the effectiveness of hindsight feedback, we conduct a quick comparative study between the \texttt{No Language} agent and the \texttt{Template Hindsight} agent in HomeGrid. The study was designed as follows:
\begin{enumerate}
    \item Both agents are driven to the same state using an expert policy.    
    \item A deliberate mistake is introduced for both agents. Three types of mistakes are designed:
    \begin{itemize}
        \item \textbf{Navigation Mistake}: The agent moves in the opposite direction compared to the expert action.
        \item \textbf{Object Pick/Drop Mistake}: The agent picks or drops an object when the expert action is to drop or pick, respectively.
        \item \textbf{Bin Manipulation Mistake}: The agent chooses the wrong action among pedal/lift/grasp to open a specific trash bin.
    \end{itemize}
    \item We use expert actions as the ground truth (GT) actions and compare the performance of both agents over 500 runs. 
\end{enumerate}
The results are shown in Appendix Table \ref{tab:hind_correction_rst}:
\begin{table}[ht]
    \resizebox{0.5\textwidth}{!}{%
    \begin{tabular}{lcc}
        \hline
        \textbf{Mistake Type} & \textbf{No Lang (\%)} & \textbf{Template Hindsight (\%)} \\ \hline
        Navigation            & 37.6 $\pm$ 0.3              & 46.2 $\pm$ 0.2                         \\ 
        Object Pick/Drop      & 37.4 $\pm$ 2.5              & 41.8 $\pm$ 1.6                         \\ 
        Bin manipulation      & 23.5 $\pm$ 1.2              & 24.8 $\pm$ 0.9                         \\ \hline
    \end{tabular}
    }
    \caption{Comparison of performance between \texttt{No Language} Agent and \texttt{Template Hindsight} Agent on different Mistake Types.}
    \label{tab:hind_correction_rst}
    \vspace{-12pt}
\end{table}
The results indicate that for the navigation and object pick/drop mistakes, hindsight feedback is highly beneficial. This is because identifying a wrong action usually directly implies the correct action for those mistakes (e.g., if "turn left" is wrong, "turn right" is correct; if "pick the object" is wrong, "drop the object" is correct). However, for the bin manipulation mistake, hindsight feedback is less helpful since the action space grows larger (pedal/lift/grasp, compared to binary opposite actions in Navigation and Object Pick/Drop), and there are no clear implications for the correct action.

\section{More results on the Messenger environment}
\label{sec:messager-res}
\begin{figure}
\centering
\includegraphics[width=0.5\textwidth]{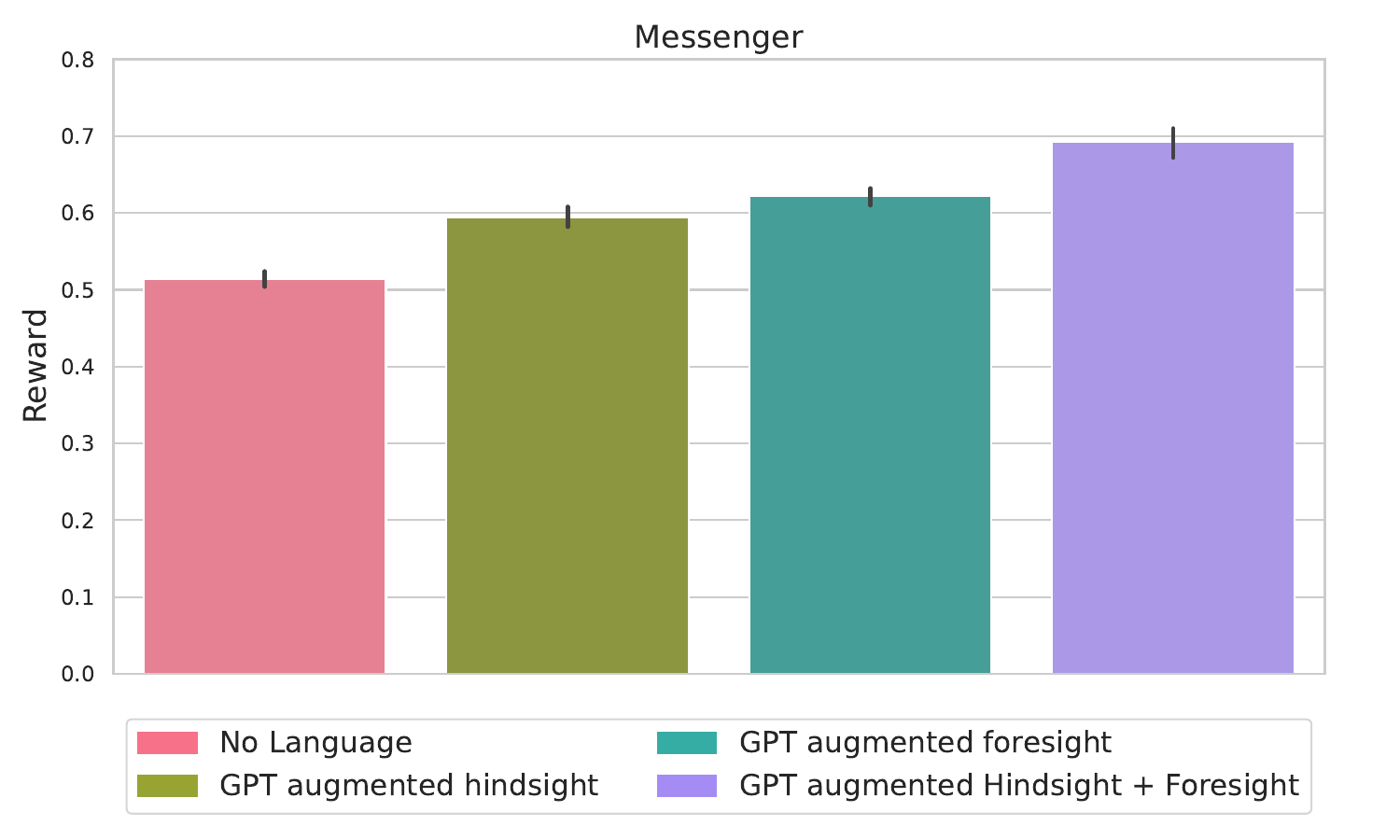}
\caption{In the Messenger environment, when trained with more diverse foresight and hindsight languages, the agents can perform better than those trained without languages. Furthermore, agents trained with more informative languages demonstrate stronger performance.}
\label{explain_messenger}
\end{figure}

In the Messenger environment, models trained with only template foresight or hindsight languages struggle to generalize to diverse languages during testing. Without exposure to diverse languages during training, these models fail to extract the learned hindsight or foresight information from mixed and diverse languages. However, Figure \ref{explain_messenger} demonstrates that models trained with more diverse hindsight or foresight languages can overcome the generalization problem, and outperform those trained without language feedback, showcasing the importance of diversity in the training languages. Furthermore, the agents trained with both hindsight and foresight information still perform the best, aligning with results in other environments.

\section{Models and Training}
\label{models_and_training}
We build our Language-Teachable Decision Transformer based on the code of the original Decision Transformer \cite{chen2021decision}. In this section, we will show our training setup and model hyperparameters for each environment.

When selecting the data size, we prioritize the efficient use of a small-scale dataset and examine the impact of language feedback within the constraints of a limited budget and scarce data, as is common in the field of robotics.

\subsection{HomeGrid}
Estimated parameter size of the models: 12.191 MB. For research question 1, we train the model with 100 trajectories. For research question 2, the pretraining stages use 6432 trajectories. The models are trained on one Nvidia RTX A6000. For research question 1, training takes 3 GPU hours. For research question 2, pretraining takes 4 GPU hours and adaptation takes 3 GPU hours. Hyperparameters shown in Appendix Table \ref{homegrid_setting}.
\subsection{ALFWorld}
Estimated parameter size of the models: 6.5 MB. For research question 1, we train the model with 1000 trajectories. For research question 2, the pretraining stages use 10000 trajectories. The models are trained in one Nvidia RTX A6000. For research question 1, training takes 3 GPU hours. For research question 2, pretraining takes 4 GPU hours and adaptation takes 3 GPU hours. Hyperparameters shown in Appendix Table \ref{ALFWorld_setting}.
\subsection{Messenger}
Estimated parameters size of the models: 289.681 MB. We train the models with 10000 data trajectories during the pretraining stage for seen tasks. The pretraining stage for seen tasks takes 5 GPU hours on one Nvidia RTX A6000. The adaptation stage for unseen tasks takes 1 GPU hour.
Hyperparameters are shown in Appendix Table \ref{messenger_setting}.
\subsection{MetaWorld}
Estimated parameters size of the models: 289.681 MB. We train the models with 20000 data trajectories during the pretraining stage for seen tasks. The pretraining stage for seen tasks takes 2.5 GPU hours on one Nvidia RTX A6000. The adaptation stage for unseen tasks takes 1 GPU hour. Hyperparameters are shown in Appendix Table \ref{metaworld_setting}.

\begin{table}[ht]
\resizebox{0.5\textwidth}{!}{%
\begin{tabular}{ll}
\hline
\textbf{Hyperparameters}  & \textbf{Value}                                \\ \hline
Number of transformer layers                    & 3                                             \\
Number of attention heads           & 1                                             \\
Embedding dimension                 & 128                                           \\
Nonlinearity function               & ReLU                                          \\
Batch size                          & 64                                            \\
Context length $K$                  & 10                                            \\
Return-to-go conditioning           & 1.5                                           \\
Dropout                             & 0.1                                           \\
Optimizer                           & AdamW                                         \\
Learning Rate                       & $1e^{-4}$                                     \\
Grad norm clip                      & 0.25                                          \\
Weight decay                        & $1e^{-4}$                                     \\
Learning rate decay                 & Linear warmup for first $1e^5$ training steps \\ \hline
\end{tabular}%
}
\caption{Hyperparameters of Language-Teachable Decision Transformer for HomeGrid experiments.}
\label{homegrid_setting}
\end{table}
\begin{table}[ht]
\resizebox{0.5\textwidth}{!}{%
\begin{tabular}{ll}
\hline
\textbf{Hyperparameters}  & \textbf{Value}                                \\ \hline
Number of transformer layers                    & 3                                             \\
Number of attention heads           & 1                                             \\
Embedding dimension                 & 128                                           \\
Nonlinearity function               & ReLU                                          \\
Batch size                          & 64                       \\
Context length $K$                  & 10                                            \\
Return-to-go conditioning           & 1.5                                           \\
Dropout                             & 0.1                                           \\
Optimizer                           & AdamW                                         \\
Learning Rate                       & $1e^{-3}$                             \\
Grad norm clip                      & 0.25                                          \\
Weight decay                        & $1e^{-4}$                                     \\
Learning rate decay                 & Consine Annealing with minimum $lr=1e^{-5}$ \\ \hline
\end{tabular}%
}
\caption{Hyperparameters of Language-Teachable Decision Transformer for ALFWorld experiments.}
\label{ALFWorld_setting}
\end{table}

\begin{table}[ht]
\resizebox{0.5\textwidth}{!}{%
\begin{tabular}{ll}
\hline
\textbf{Hyperparameters}  & \textbf{Value}                                \\ \hline
Number of transformer layers                    & 5                                             \\
Number of attention heads           & 2                                             \\
Embedding dimension                 & 128                                           \\
Nonlinearity function               & ReLU                                          \\
Batch size                          & 128 for pertaining and 1 for adaptation                                         \\
Context length $K$                  & 10                                            \\
Return-to-go conditioning           & 1.5                                           \\
Dropout                             & 0.1                                           \\
Optimizer                           & AdamW                                         \\
Learning Rate                       &  $1e^{-3}$ for pretraining and $1e^{-4}$ for adaptation                                             \\
Grad norm clip                      & 0.25                                          \\
Weight decay                        & $1e^{-4}$                                     \\
Learning rate decay                 & Linear warmup for first $1e^5$ training steps \\ \hline
\end{tabular}%
}
\caption{Hyperparameters of Language-Teachable Decision Transformer for Messenger experiments.}
\label{messenger_setting}
\end{table}

\begin{table}[ht]
\resizebox{0.5\textwidth}{!}{%
\begin{tabular}{ll}
\hline
\textbf{Hyperparameters}  & \textbf{Value}                                \\ \hline
Number of transformer layers                    & 5                                            \\
Number of attention heads           & 2                                             \\
Embedding dimension                 & 256                                           \\
Nonlinearity function               & ReLU                                          \\
Batch size                          & 128 for pertaining and 5 for adaptation                                                                \\
Context length $K$                  & 12                                            \\
Return-to-go conditioning           & 20                                           \\
Return scale                        & 10                                            \\
Dropout                             & 0.1                                           \\
Optimizer                           & AdamW                                         \\
Learning Rate                       & $1e^{-5}$ for pertaining and $1e^{-6}$ for adaptation                                   \\
Weight decay                        & $1e^{-4}$                                     \\
Learning rate decay                 & Linear warmup for first $1e^5$ training steps \\ \hline
\end{tabular}%
}
\caption{Hyperparameters of Language-Teachable Decision Transformer for MetaWorld experiments.}
\label{metaworld_setting}
\end{table}

\section{Examples for Language Feedback in Evaluation}
\label{Appendix:Language Feedback in Evaluation}
\begin{figure}[ht]
    \centering
    \includegraphics[width=0.5\textwidth]{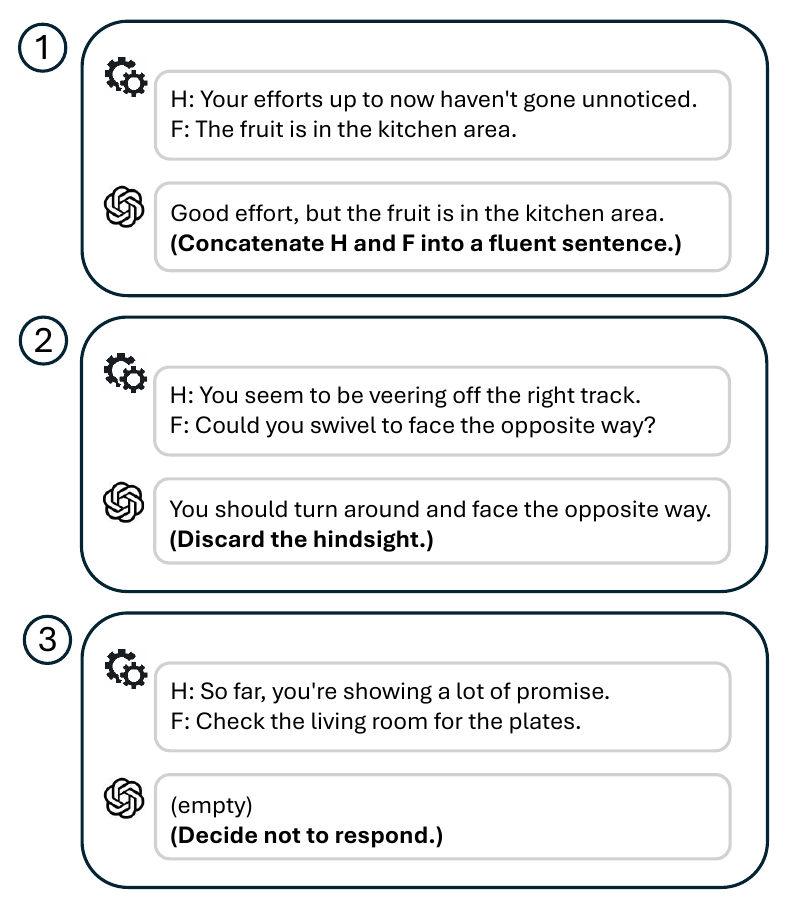}
    \caption{Examples for language feedback generated by online GPT in evaluation.}
    \label{fig:appendixD}
\end{figure}

As discussed in section \ref{sec:eval_lang_feedback}, we feed template hindsight ($l^{hind}$) and template foresight ($l^{fore}$) into an online GPT to generate language feedback as a proxy for real-world human feedback, which can be further extended into multi-turn human-machine dialogue systems in task-oriented settings \cite{he2022space, he2022unified, he2022galaxy}. In Figure \ref{fig:appendixD}, we demonstrate three examples of the GPT outcome. In example 1, we find GPT can concatenate both hindsight and foresight and integrate them into a new fluent sentence. In the second example, we observe that GPT decides to discard the hindsight part and provides only foresight as the outcome. In example 3, GPT chooses not to respond when it thinks the current agent doesn't need help. 

\end{document}